# Dynamic Occupancy Grids for Object Detection: A Radar-Centric Approach

Max Peter Ronecker[1,3], Markus Schratter[2], Lukas Kuschnig[2] and Daniel Watzenig[2,3]

*Abstract*— Dynamic Occupancy Grid Mapping is a technique used to generate a local map of the environment, containing both static and dynamic information. Typically, these maps are primarily generated using lidar measurements. However, with improvements in radar sensing, resulting in better accuracy and higher resolution, radar is emerging as a viable alternative to lidar as the primary sensor for mapping. In this paper, we propose a radar-centric dynamic occupancy grid mapping algorithm with adaptations to the state computation, inverse sensor model, and field-of-view computation tailored to the specifics of radar measurements. We extensively evaluate our approach with real data to demonstrate its effectiveness and establish the first benchmark for radar-based dynamic occupancy grid mapping using the publicly available Radarscenes dataset.

## I. INTRODUCTION

Perceiving the environment is a critical task for highly automated vehicles to ensure both safety and optimal performance. To achieve this, a variety of algorithms have been developed to process the data collected by sensors and gain a comprehensive understanding of the surroundings. Traditionally, this process has been divided into two key aspects: the detection and tracking of dynamic objects such as cars and pedestrians, and the representation of the static environment using occupancy grid maps [1]. However, in highly dynamic scenarios, which automated vehicles frequently encounter during their operation, occupancy grids have difficulties in accurately mapping moving objects causing fragments and elongation. To address this issue, the concept of Dynamic Occupancy Grid Maps (DOGM) was introduced [2] and subsequently improved by combining it with a particle filter [3]. Modern Dynamic Occupancy Grid Maps (DOGMs) are used to keep track of both stationary and moving objects in an environment. They do this by using a grid to represent the static parts and a particle filter to follow the dynamic elements within the grid map. The key benefit of DOGMs is that they can handle both the unchanging and changing aspects of the environment together. This provides a consistent foundation for tasks like planning and perception. Many variations of DOGMs exist [4]–[12], each with its own way of handling tracking or grid information. However, most of these variants primarily rely on lidar sensors for data, using radar to improve certain aspects like convergence and speed estimation [6]. Lidar sensors are a popular choice because they provide the detailed point cloud data needed to create a comprehensive grid map. In recent years, there have been improvements in both the detection density and resolution of radar sensors, potentially enabling radar-only DOGMs. Research has been conducted to develop radar-based DOGMs [13], [14], but these have only been tested in short scenarios and with minor modifications to the lidar-based algorithms. However, radar has distinct advantages, including range rate measurements and resilience in various weather and lighting conditions. Unfortunately, existing algorithms neither exploit these advantages nor effectively mitigate issues like data sparsity and noise. This paper explores different ways to adapt current algorithms in order to create effective radar-based DOGMs.

[1]SETLabs Research GmbH, 80687 Munich, Germany first.last@setlabs.de
[2] Virtual Vehicle Research GmbH, 8010 Graz, Austria first.last@v2c2.at
[3] Graz University of Technology, Institute of Automation and Control, 8010 Graz, Austria first.last@tugraz.at

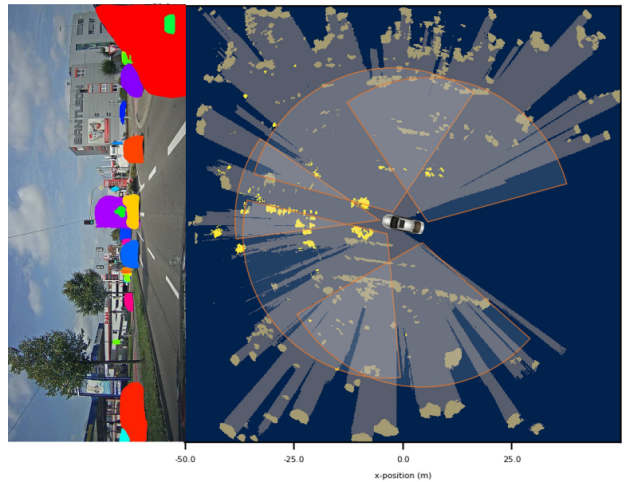

Fig. 1: Example of a DOGM using radar data from Radarscenes [15]. The cell coloring indicates which states have the highest probability (yellow=*dynamic*,brown=*static*,gray=*free* and blue=*unknown*).

**Contributions**

In this paper, we present several modifications to the existing lidar-based DOGM algorithms in order to leverage the advancements in modern radar sensors and validate them extensively using real data.

1) We propose a cell-state classification scheme based on range rate that allows effective separation of static and dynamic objects and significantly reduces the number of particles needed for computing the dynamics within the grids.
2) We integrate several modifications into the Inverse Sensor Model (ISM) used to generate the DOGM to account for the characteristics of radar sensors.

3) We extensively validate the effectiveness of the modifications and compare their performance to that of other DOGMs and radar-based object detectors. To the best of our knowledge, we are the first to assess the performance of Radar-DOGMs, both qualitatively and quantitatively, in small test scenarios and public datasets.

## II. RELATED WORK

The original Bayesian Occupancy Filter (BOF), as described in earlier work [2], utilized a four-dimensional grid, which had significant computational demands due to the velocity computation. To address these challenges, a more efficient version of a BOF was introduced in [3]. This improved approach combined a grid representation with a particle filter to estimate both velocity and occupancy distribution within the grid, helping with the identification of dynamic cells. This methodology has been adopted and refined in subsequent works [4], [5]. Furthermore, the fusion of lidar and radar sensor measurements, can enhance the filter's performance [6]. In addition, a Dempster-Shafer [16]–[18] representation of a DOGM was introduced [6], [8], which was further enhanced [7] and expanded to incorporate object detection and shape estimation tasks [9]–[12]. Deep neural networks have also been proposed for learning dynamic occupancy grid mapping end-to-end [19], [20]. However, most the research in the field of (dynamic) occupancy grids has largely focused on using lidar sensors as the main source of information, mostly due to the dense data they provide. Radar, up to this point, has been only utilized to support velocity estimation speed up convergence. There have been attempts to adapt existing algorithms for radar-only use. An extension of the work presented in [6], [7] was introduced in [13], which aims to exclusively utilize radar data and combine it with a clustering algorithm to provide object-level information. Moreover, a modification to the weight calculations introduced in [4], [5] has been proposed in [14]. Nevertheless, the core algorithmic structure remains unaltered, and the performance of DOGMs, whether radar or lidar-based, is primarily assessed in short and simple scenarios. Additionally, some research has explored radar-centric methods for traditional static occupancy grids. Factors specific to radar, such as signal amplitude and field of view, have been taken into account in the standard Inverse Sensor Model (ISM) [21]–[23]. There have also been efforts to learn improved radar-ISMs through a combination of deep learning and lidar data [24], [25].

## III. METHODOLOGY

### A. Environment representation

The area surrounding the vehicle is depicted using a grid map centered on the ego-vehicle's position. In this grid, each cell stores probabilities for one of the four states: "free" (unoccupied), "static" (occupied by a stationary object), "dynamic" (occupied by a moving object) and "unknown" (not yet observed). Additionally, particles described by the state vector $\mathbf{x} = [x, y, v_x, v_y, weight]^T$ are used to help track dynamic occupancy. Each particle is associated with a specific cell based on its position. Fig. 2 provides a visual representation of this concept.

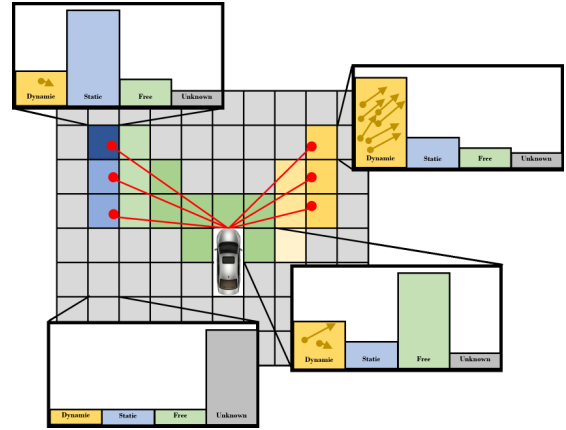

Fig. 2: Environment representation using a dynamic occupancy grid. Each cell contains probabilities for the four states: *unknown, free, static* and *dynamic*. The kinematics are expressed through a set of particles [14].

### B. Radar-centric Dynamic Occupancy Grid Mapping

In the following, our radar-centric DOGM algorithm to effectively use radar data and the reasoning behind the modifications are introduced. The algorithm is based on [4], [5] and a full description of the original algorithm can be found in the mentioned references. The introduced steps are executed for each new scan of radar measurements.

*1) Determine measurement cells:* In the initial step, we identify the cells influenced by the latest measurements. The objective is to determine which cells might be classified as free, occupied, or remain unknown. This constitutes the first part of the Radar Inverse Sensor Model (ISM). A standard solution for lidar data involves ray-casting using Bresenham's line algorithm [26], which assumes straight, narrow rays, that align with the properties of laser rays. However, this assumption does not hold for radar measurements, as radar waves exhibit relatively high angular uncertainty and can penetrate materials to varying degrees. Consequently, this leads to an ambiguous line-of-sight geometry, where free space can overwrite occupancy [27].

Additionally, radar sensors provide significantly fewer measurements compared to lidar sensors and lack the mechanical structure required to determine ray traces for filling empty areas. Consequently, large portions of the map remain unknown. To address these limitations, we combine the Inverse Sensor Model (ISM) from [21], which considers high angular uncertainty and wave propagation effects, with the implicit free-space computation described in [22]. This combination enables us to create a much more comprehensive and precise map. Using our Radar-FOV-ISM $\Phi_{\text{radar}}$, we produce $C_{\text{meas}}$, detailing unknown ($C_{\text{unk}}$), free ($C_{\text{free}}$), and occupied ($C_{\text{occ}}$) cells as index pairs $(i, j)$, e.g., $C_{\text{unk}} = (i_1, j_1), \cdots$, of the 2D grid map.

$$X_{\text{meas}} = \{x_{\text{meas}}^1, x_{\text{meas}}^2, \cdots, x_{\text{meas}}^n\} \qquad (1)$$

$$C_{\text{meas}} = \Phi_{\text{radar}}(X_{\text{meas}}) \quad (2)$$

$$C_{\text{meas}} = \{C_{\text{unk}}, C_{\text{free}}, C_{\text{occ}}\} \quad (3)$$

A comparison of the standard ray-casting model and the combined Radar-FOV-ISM is shown in Fig. 3a.

*2) Cell state computation:* In DOGM algorithms, there exist four possible states: *unknown*, *free*, *static*, and *dynamic*. For each cell, the probability of all four states is stored. A fundamental assumption across all frameworks is that freespace is always static since there is nothing that can move. To determine the static and dynamic occupancy probabilities, we directly utilize the range rate of the measurement. This approach differs from all lidar-focused approaches [4]–[12], which rely on the average velocity of particles created in every occupied cell.

Each radar measurement $x_{\text{meas}}^n$ includes position coordinates in the map frame ($x_{\text{map}}, y_{\text{map}}$), ego-motion-compensated range rate ($v_r$), and radar cross-section ($rcs$). For each measurement, the subsequent calculations are executed.

$$\mathbf{x}_{\text{meas}}^n = [x_{\text{map}}, y_{\text{map}}, v_r, rcs]^T \quad (4)$$

In a first step free space, which has no static or dynamic property can be calculated based on the distance to the measurements [14]. For each cell of $C_{\text{free}}$ and $C_{\text{occ}}$ the respective center coordinates ($x_c, y_c$) and the distance $d_c$ to the measurement is calculated.

$$d_c = \left\| \begin{bmatrix} x_{\text{map}} \\ y_{\text{map}} \end{bmatrix} - \begin{bmatrix} x_c \\ y_c \end{bmatrix} \right\|_2 \quad (5)$$

This distance is then used to determine the free space probability $P(free)$ with a $\sigma_d$ of 1.0.

$$f_d(d_c) = \frac{1}{\sigma_d \sqrt{2\pi}} \exp\left(-\frac{1}{2}\left(\frac{d_c}{\sigma_d}\right)^2\right) \quad (6)$$

$$P_{\text{free}} = 1 - f_d(d_c) \quad (7)$$

The probability $P_{\text{meas}}(unk)$ for the unknown state is calculated in the same way as the freespace but for the cells $C_{\text{unk}}$.

To determine the static and dynamic probabilities, we operate under the assumption that a non-zero range rate can indicate whether an object is in motion, thereby allowing us to distinguish between static and dynamic cells. However, it's important to note that this assumption doesn't hold true in all cases. Slow-moving objects like pedestrians or objects moving perpendicular to the sensor axis can exhibit very low range rates. Nevertheless, such cases are relatively rare, as illustrated in Fig. 4. More than 90% of larger objects (cars, large vehicles, two-wheelers) and approximately 70% of pedestrians (groups) have radar detections with a range rate exceeding 0.5 m/s.

Conversely, static cells can be clearly identified, rendering it unnecessary to employ particles for all occupied cells. This method significantly reduces the number of required particles and, consequently, computation time, as demonstrated later. Utilizing the range rate $v_r$ leads to the following equations for determining the state probabilities for all cells in $C_{\text{occ}}$. In the equations, *prior* represents the cell's previous probability of the respective state.

$$P_{\text{dyn}} = f_d(d) \times P(v_r \neq 0) + (1 - f_d(d)) \times prior(dyn) \quad (8)$$

$$P_{\text{static}} = f_d(d) \times P(v_r = 0) + (1 - f_d(d)) \times prior(static) \quad (9)$$

The distance function is used to reduce the influence of a measurement as the distance increases. Additionally, we suggest utilizing the normalized radar-cross section (RCS) $rcs'$ to weigh the contribution of a measurement. Measurements with a low RCS are more likely to be noise, while higher RCS values indicate a more reliable measurement. The RCS is normalized between 0 and 1 for each measurement scan using all the rcs values $RCS_{\text{meas}}$ of all measurements $\mathbf{x}_{\text{meas}}^n$ of one scan.

$$RCS_{\text{meas}} = \{rcs^1, rcs^2, \cdots, rcs^n\} \quad (10)$$

$$rcs' = \frac{rcs - \min(RCS_{\text{meas}})}{\max(RCS_{\text{meas}}) - \min(RCS_{\text{meas}})} \quad (11)$$

Applying this the measurement state probabilities $P_{\text{meas}}(c)$ of one cell $c$ can be described as the following:

$$\begin{bmatrix} P_{\text{meas}}(unk) \\ P_{\text{meas}}(free) \\ P_{\text{meas}}(static) \\ P_{\text{meas}}(dyn) \end{bmatrix} = rcs' \begin{bmatrix} P_{\text{unk}} \\ P_{\text{free}} \\ P_{\text{static}} \\ P_{\text{dyn}} \end{bmatrix} + (1 - rcs') \begin{bmatrix} 0.5 \\ 0.5 \\ 0.5 \\ 0.5 \end{bmatrix} \quad (12)$$

Finally, all the calculated probabilities are stored in the form of a measurement grid and are used to update the existing grid using a Bayesian formulation.

*3) State correction:* As shown in Fig. 4 the assumption that moving objects have a non-zero range rate doesn't hold true for all cases. Specifically, slow-moving objects or those moving perpendicular to the sensors may have a low range rate despite being in motion. In order to deal with this we propose two methods that correct cell states of the measurement grid or within the dynamic grid map.

**Measurement Grid Correction:** The assumption is that a highly dynamic cell (indicated by high velocity) cannot abruptly become static. This situation may occur, for example, in the case of crossing vehicles when the range rate approaches zero. To tackle this, we suggest updating the measurement grid based on the existing accumulated DOGM. This means that a cell in the measurement grid, which has a high velocity in the existing DOGM, has a reduced probability of being static, even if the latest measurement suggests otherwise (see Fig. 3b). This can be expressed using the following equation for each cell:

$$P_{\text{meas},c} = \begin{bmatrix} P_c(unk) \\ P_c(free) \\ P_c(static) \\ P_c(dyn) \end{bmatrix} \quad (13)$$

$$P_{\text{corr.},c} = \begin{pmatrix} 1 & 0 & 0 & 0 \\ 0 & 1 & 0 & 0 \\ 0 & 0 & 1 - s_l P(v \neq 0) & d_l P(v = 0) \\ 0 & 0 & s_l P(v \neq 0) & 1 - d_l P(v = 0) \end{pmatrix} P_{\text{meas},c} \quad (14)$$

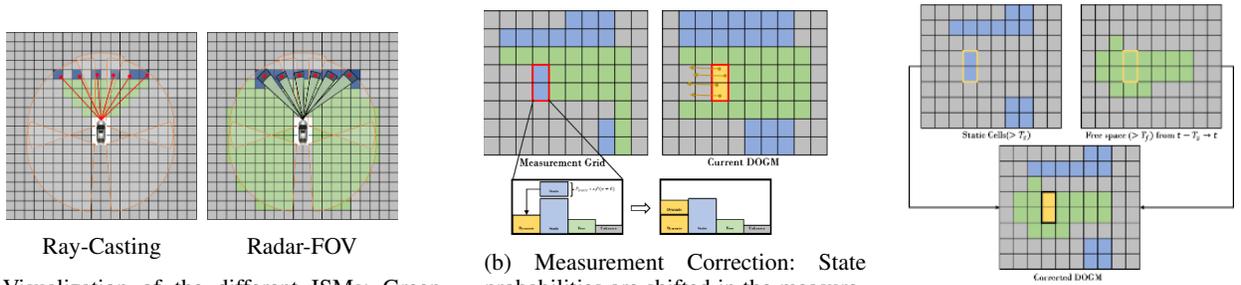

| Ray-Casting | Radar-FOV | | | | |
|---|---|---|---|---|---|
| (a) Visualization of the different ISMs: Green areas represent $C_{\text{free}}$, blue is $C_{\text{occ}}$ and gray $C_{\text{unk}}$. It shows how the Radar-ISM create a more complete map. | | (b) Measurement Correction: State probabilities are shifted in the measurement grid based on the velocity in the current DOGM. The red area highlights false static cells. | | (c) False Static Detection: Wrongly classified static cells are detected if they appear in high-confidence freespace. | |

Fig. 3: Overview of the methods used to adapt the DOGM algorithms for radar

The probabilities $P(v=0)$ and $P(v \neq 0)$ are calculated based on the mean particle velocity in the respective cell of the DOGM. The parameters $s_l$ and $d_l$ fall within the range of 0 to 1 and determine the weighting between the influence of the latest measurement and the existing grid map. They are determined empirically, and for this paper, $s_l$ and $d_l$ are both set to 0.5. This entire correction process can be efficiently implemented using stacked matrices.

**False Static Detection**: One advantage of using DOGMs is that none of the measurements are discarded; they are only at risk of being misclassified. This risk is especially high for slow-moving objects, such as pedestrians, which can be falsely classified as static due to their low walking speed. Nevertheless, these false static objects remain a part of the map and subsequently can still be detected. To correct this, we adapt the Dynablox algorithm, originally proposed in [28], for radar applications. The basic detection mechanism involves identifying objects that appear in highly confident free space areas, as the only way for them to be present there is through motion. Therefore, they need to be dynamic objects. Instead of using a time threshold, as in the original paper, we count the number of cycles during which a cell has consistently been classified as static or free space (see Fig 3c). If a cell is classified as static for more than $T_{\text{static}}$ cycles (e.g., 4 cycles) and was previously classified as free space for more than $T_{\text{free}}$ cycles, it is transformed into a dynamic state. This requires storing the history of high-confidence free space for $T_{\text{static}}$ cycles. Tuning the parameters $T_{\text{static}}$ and $T_{\text{free}}$ determines the sensitivity of the detection algorithm. Overall, this approach is primarily effective for slow-moving objects that stay in a cell for an extended duration. But it aligns with the objective of detecting objects with low speed, as faster-moving objects can be identified using the primary classification method. Both approaches can be implemented efficiently and help to compensate for radar-specific issues without relying on a high number of particles.

*4) Particle creation, weight computation and Resampling:* After determining the probability of each state per cell, particles are created in the newly identified dynamic cells. A cell is classified as dynamic if its corresponding probability is the highest (determined by a maximum operation). These particles become part of the existing pool of particles. A particle is characterized by the following state:

$$\mathbf{x}_p = [x_p, y_p, v_{p,x}, v_{p,y}, w_t]^T \quad (15)$$

The particle velocity is defined in the global frame, while the position coordinates are in the grid map frame. The weight indicates the likelihood of the particle representing the tracked object based on the measurements. For the weight computation, we use the range rate as proposed in [6].

$$w_t = \underbrace{f_d(d_p) f_r(\mathbf{r}_p)}_{\text{update}} \underbrace{(1 - f_d(d_p))(1-\varepsilon) w_{t-1}}_{\text{prior}} \quad (16)$$

In Equation 15, $d_p$ represents the distance between the particle and the nearest measurement, which is utilized to attenuate the measurement's impact based on the distance. The function $f_r(r_p)$ is a probability density function that compares the particle's range rate $r_p$ with the range rate of the measurement. While this provides only partial information, when combined with the fact that only particles with the appropriate velocity can be consistently tracked, it is sufficient for obtaining a reliable velocity estimate of the object.

Furthermore, to narrow down the potential range of velocities, particles are exclusively sampled with a velocity that matches the range rate of the nearest measurement and is constrained by a maximum velocity [29]. Subsequently, based on their weight, the particles undergo resampling as described in [4].

*5) Normalization and Occupancy Prediction:* In the final step, all state probabilities and particle weights must be correctly normalized to ensure accurate probabilities. Additionally, the dynamic state probability is distributed among the particle weights in the respective cell, proportionate to their weight. Subsequently, the particles move according to a linear motion model, allowing for the prediction of dynamic occupancy for the next time step. We employ the following state transition matrix based on [5].

$$S_t = \begin{pmatrix} 1 & 0.1 & 0.1 & 0.05 \\ 0 & 0.9 & 0 & 0 \\ 0 & 0 & 1 - P(v \neq 0) & P(v = 0) \\ 0 & 0 & P(v \neq 0) & 1 - P(v = 0) \end{pmatrix} \quad (17)$$

The probabilities $P(v \neq 0)$ and $P(v = 0)$ are limited with respect to the part that is transitioned to the unknown state.

## IV. EVALUATION

### A. Evaluation data

To assess the performance of our radar-centric DOGM, we conduct both qualitative and quantitative analyses. For the quantitative evaluation, we recorded two simple scenarios with our test vehicle: one with a vehicle crossing (30 kph) in front of the ego vehicle and another with a pedestrian crossing. The radars used include a smartmicro UMRR-11 in the front and four UMRR-96 at each corner of the vehicle. In these cases, we aim to demonstrate the effectiveness of our proposed methods, especially in scenarios where range-rate measurements could potentially lead to false static object classification. We generate a list of dynamic objects by clustering the particles inside the dynamic cells using the HDB-SCAN [30] algorithm and compare them with the center of the bounding boxes. In all cases, the map resolution is 0.2m and size depends on the evaluation area. The groundtruth for the qualitative analysis is generated using a state-of-the-art 3D lidar-object detector and tracker of the Autoware stack. Additionally, we compare our results with the original HSBOF combined with the new state calculation (HSBOF-RS) but without the other radar-specific modifications. For the quantitative analysis and to compare our DOGMs with other state-of-the-art radar-object detection frameworks, we utilize the validation set of the Radarscenes dataset [15]. This dataset provides point-wise labels exclusively for dynamic objects, aligning well with DOGMs, which primarily tracks moving objects (dynamic cells). However, it does not provide bounding boxes. Consequently, we perform a preprocessing step to generate bounding boxes using the method outlined in [31], [32]. We will also use their proposed object detectors, one being graph neural net based and the other two relying on a YOLO and point pillar architecture as benchmarks. Unfortunately, there are no other benchmarked Radar-DOGM algorithms, so we only compare them to the referenced machine-learning-based radar object detectors. The metric used for comparison in the radarscenes dataset is the mean Average Precision (mAP) as defined in [31]. However, since we don't generate bounding boxes, we use the mAP calculation from the nuscenes dataset [33], which is distance-based instead of intersection-over-union. As confidence metric for generating the precision-recall curves, we use a combination of particle age and weight. We acknowledge that this comparison is not perfect due to the comparison with machine-learning solutions and differences in metrics. Nevertheless, given the lack of alternatives, this approach seems to be the most suitable option for evaluating the performance of our radar-centric DOGM.

### B. Qualitative Evaluation

Table I displays the results for the two challenging tracking scenarios. It is evident that our approach outperforms the original algorithm across all metrics. Particularly noteworthy is the higher recall. For instance, the pedestrian was not detectable due to its low speed in the basic radar-state DOGM, but in the improved DOGM, it can be tracked consistently and with high accuracy.

On the other hand, it can be observed that, although improved, the precision remains relatively low. In this scenario where only one vehicle or pedestrians need to be tracked, this value should ideally be higher. We have identified two main reasons for this. First, the clustering algorithm doesn't work perfectly, resulting in the point cloud being fragmented into several objects. Second, the Radar-Dynablox occasionally generates false dynamic cells, impacting the precision. Another noteworthy fact is that the number of particles is with under 10000 particles relatively low. Based on the information provided in the Lidar-DOGM publications [4], [7], the necessary number of particles with the original algorithms is at least 10 times higher.

The difference in performance can also be observed in Fig. 5. It illustrates that with the original algorithm, for both scenarios, the tracking is interrupted in the middle or not detectable at all, which is also a cause for the low recall. This is because the object becomes perpendicular to the sensor axis, and the range rate approaches close to 0. Hence, it's classified as static. This effect is no longer visible in the radar-centric approach. Thanks to the correction methods and Radar-Dynablox, we are able to consistently track even slow-moving objects with high accuracy, as demonstrated by the low velocity error.

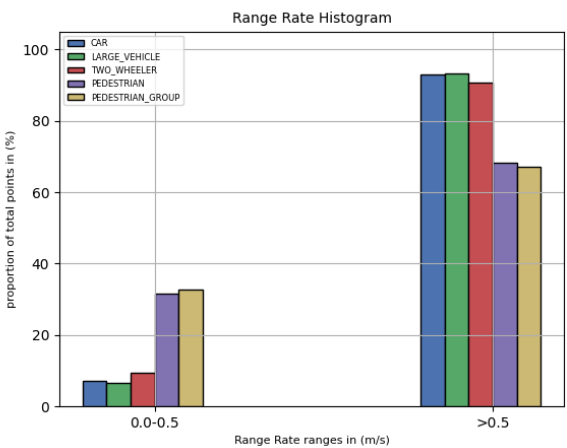

Fig. 4: Evaluation of the range rate per object class in the Radarscenes dataset [15]: In the dataset only dynamic objects are labeled, and it's evident that the majority of objects have a range rate exceeding 0.5 m/s, enabling their classification as dynamic. However, there are exceptions, particularly in the case of pedestrian classes.

TABLE I: Results on the recorded scenarios

| Method | $\Delta x$ (m) | $\Delta v$ (m/s) | recall(%) | precision(%) | N.P. |
|---|---|---|---|---|---|
| **Crossing Vehicle** | | | | | |
| HSBOF-RS | 1.3 | 2.3 | 61.0 | 12.0 | 1164 |
| **Ours** | 1.2 | 0.5 | 97.0 | 74.0 | 1889 |
| **Crossing Pedestrian** | | | | | |
| HSBOF-RS | 0.67 | 3.02 | 3.2 | 1.0 | 2145 |
| **Ours** | 1.0 | 0.14 | 83.0 | 36.0 | 7362 |

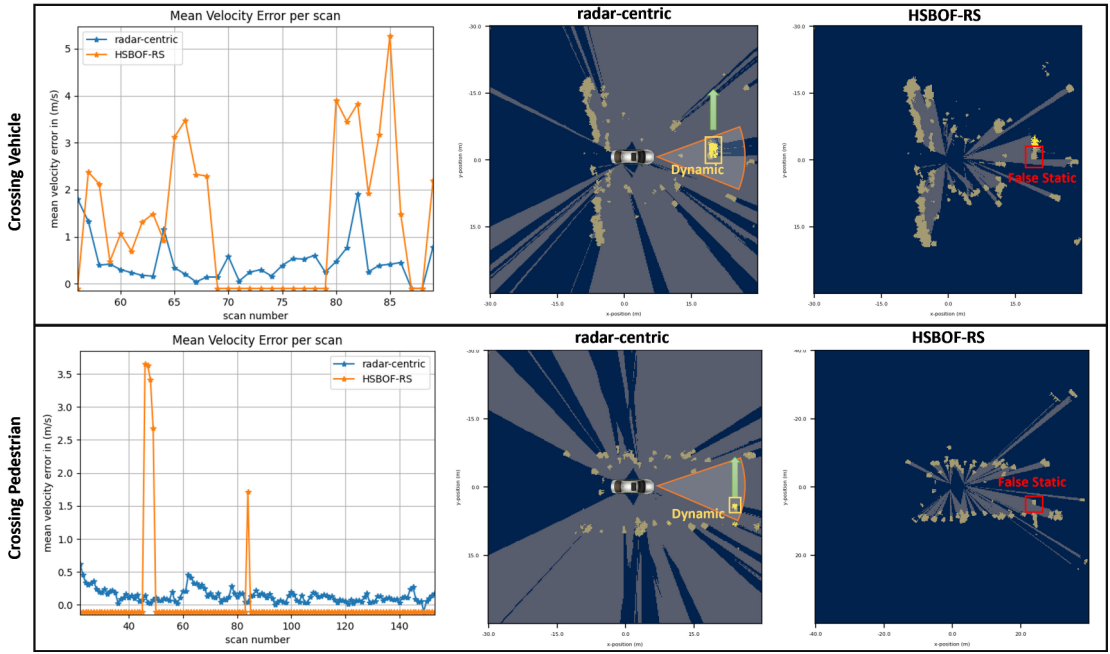

Fig. 5: Qualitative comparison: The velocity error of the Radar-centric and HSBOF-RS approach is shown for both scenarios. It shows the large gaps in the tracking as well as the significantly better velocity estimate of the radar-centric approach. In addition an example of a respective DOGM is shown visualizing how a false static area has been correctly classified as dynamic.

## C. Quantitative Evaluation

The radar-centric DOGM performance is compared to the HSBOF-RS and deep-learning-based object detectors in [32] and [31]. The results are summarized in Table II. Additionally, we provide the precision and recall of our radar-centric DOGM to better understand the performance. Unfortunately, these metrics are not available for other benchmarks.

The results indicate that our radar-centric DOGM does not outperform deep-learning-based solutions but shows in some aspects (mAP) comparable performance to the PointPillar detector [31], and even outperforms it in the pedestrian classes. This is a noteworthy achievement considering that the radar-centric DOGM does not rely on machine learning, and dynamic object detection is only one aspect of the information it can provide. Furthermore, it is significantly improved compared to the basic HSBOF-RS. Interestingly, it appears that, for all classes, performance is more constrained by precision rather than recall, although both metrics require improvement. Furthermore, the large vehicle class, which one might expect to perform best due to its size, actually has the lowest detection scores. This is likely due to the fact that larger objects are often fragmented, particularly affecting the precision metric. Enhancing precision involves improving the clustering algorithm and fine-tuning the RadarDynablox, thereby reducing false positives. Addressing recall requires a more sensitive dynamic classification, achieved by integrating elements of the particle-based classification from the traditional DOGM algorithm into our radar-centric approach. In cases of low recall within DOGM, the issue isn't object absence but misclassification as static. Machine learning could assist in solving this classification challenge.

TABLE II: Results on the Radarscenes validation dataset. HSBOF-RS/Ours are NuScenes mAP.

| Radarscenes (val) | | | | | | |
|---|---|---|---|---|---|---|
| *Method* | $AP_{car}$ | $AP_{large}$ | $AP_{tw}$ | $AP_{ped}$ | $AP_{grp}$ | *mAP* |
| HSBOF-RS | 30.1 | 24.9 | 10.7 | 5.1 | 7.3 | 15.6 |
| PointPillars[31] | 54.2 | 53.7 | 38.8 | 11.2 | 22.5 | 36.1 |
| YOLOv3[31] | 70.2 | 61.9 | 57.4 | 34.4 | 55.7 | 55.9 |
| RadarGNN[32] | 72.0 | 70.1 | 66.6 | 34.0 | 58.3 | 60.2 |
| **Ours** | 38.7 | 24.1 | 35.7 | 30.6 | 34.5 | 32.7 |
| Recall (R) and Precision (P) | | | | | | |
| *Method* | $R_{car}$ | $R_{large}$ | $R_{tw}$ | $R_{ped}$ | $R_{grp}$ | *Recall* |
| **Ours** | 47.3 | 32.2 | 47.7 | 39.9 | 45.0 | 42.4 |
| Method | $P_{car}$ | $P_{large}$ | $P_{tw}$ | $P_{ped}$ | $P_{grp}$ | *Prec.* |
| **Ours** | 16.4 | 5.4 | 27.6 | 14.1 | 17.0 | 16.1 |

Overall, we did not expect to surpass state-of-the-art deep-learning-based radar object detectors. Instead, our objective was to assess the performance of a radar-based DOGM and get an understanding of its shortcomings and strengths. It can serve as a first benchmark and provide insights for future research directions and improvements.

## V. CONCLUSION

In conclusion, this paper presents a radar-centric DOGM algorithm and the first comprehensive analysis of radar-based DOGMs, demonstrating their potential for object detection. Our approach outperforms traditional DOGM algorithms when radar is utilized as input, showcasing the efficacy of our radar-centric approach. For future work, we aim to further enhance classification accuracy by integrating machine learning techniques. Additionally, we see potential in utilizing DOGMs as input for object detection algorithms, for example, in a BEV-Fusion framework [34].

## References

[1] A. Elfes, "Using occupancy grids for mobile robot perception and navigation," *Computer*, vol. 22, no. 6, pp. 46–57, 1989.

[2] C. Coué, C. Pradalier, C. Laugier, T. Fraichard, and P. Bessière, "Bayesian occupancy filtering for multitarget tracking: An automotive application," *The International Journal of Robotics Research*, vol. 25, no. 1, pp. 19–30, 2006. [Online]. Available: https://doi.org/10.1177/0278364906061158

[3] R. Danescu, F. Oniga, and S. Nedevschi, "Modeling and tracking the driving environment with a particle-based occupancy grid," *IEEE Transactions on Intelligent Transportation Systems*, vol. 12, no. 4, pp. 1331–1342, 2011.

[4] A. Nègre, L. Rummelhard, and C. Laugier, "Hybrid sampling bayesian occupancy filter," in *2014 IEEE Intelligent Vehicles Symposium Proceedings*, 2014, pp. 1307–1312.

[5] L. Rummelhard, A. Nègre, and C. Laugier, "Conditional monte carlo dense occupancy tracker," in *2015 IEEE 18th International Conference on Intelligent Transportation Systems*, 2015, pp. 2485–2490.

[6] D. Nuss, T. Yuan, G. Krehl, M. Stuebler, S. Reuter, and K. Dietmayer, "Fusion of laser and radar sensor data with a sequential monte carlo bayesian occupancy filter," in *2015 IEEE Intelligent Vehicles Symposium (IV)*, 2015, pp. 1074–1081.

[7] D. Nuss, S. Reuter, M. Thom, T. Yuan, G. Krehl, M. Maile, A. Gern, and K. Dietmayer, "A random finite set approach for dynamic occupancy grid maps with real-time application," *The International Journal of Robotics Research*, vol. 37, no. 8, pp. 841–866, 2018. [Online]. Available: https://doi.org/10.1177/0278364918775523

[8] G. Tanzmeister, J. Thomas, D. Wollherr, and M. Buss, "Grid-based mapping and tracking in dynamic environments using a uniform evidential environment representation," in *2014 IEEE International Conference on Robotics and Automation (ICRA)*, 2014, pp. 6090–6095.

[9] G. Tanzmeister and D. Wollherr, "Evidential grid-based tracking and mapping," *IEEE Transactions on Intelligent Transportation Systems*, vol. 18, no. 6, pp. 1454–1467, 2017.

[10] S. Steyer, G. Tanzmeister, and D. Wollherr, "Object tracking based on evidential dynamic occupancy grids in urban environments," in *2017 IEEE Intelligent Vehicles Symposium (IV)*, 2017, pp. 1064–1070.

[11] ——, "Grid-based environment estimation using evidential mapping and particle tracking," *IEEE Transactions on Intelligent Vehicles*, vol. 3, no. 3, pp. 384–396, 2018.

[12] S. Steyer, C. Lenk, D. Kellner, G. Tanzmeister, and D. Wollherr, "Grid-based object tracking with nonlinear dynamic state and shape estimation," *IEEE Transactions on Intelligent Transportation Systems*, vol. 21, no. 7, pp. 2874–2893, 2020.

[13] C. Diehl, E. Feicho, A. Schwambach, T. Dammeier, E. Mares, and T. Bertram, "Radar-based dynamic occupancy grid mapping and object detection," in *2020 IEEE 23rd International Conference on Intelligent Transportation Systems (ITSC)*, 2020, pp. 1–6.

[14] M. P. Ronecker, M. Stolz, and D. Watzenig, "Dual-weight particle filter for radar-based dynamic bayesian grid maps," in *2023 IEEE International Conference on Mobility, Operations, Services and Technologies (MOST)*, 2023, pp. 187–192.

[15] O. Schumann, M. Hahn, N. Scheiner, F. Weishaupt, J. Tilly, J. Dickmann, and C. Wöhler, "RadarScenes: A Real-World Radar Point Cloud Data Set for Automotive Applications," Mar. 2021. [Online]. Available: https://doi.org/10.5281/zenodo.4559821

[16] G. Shafer, *A Mathematical Theory of Evidence*. Princeton University Press, 1976. [Online]. Available: http://www.jstor.org/stable/j.ctv10vm1qb

[17] P. Smets, "The combination of evidence in the transferable belief model," *IEEE Transactions on Pattern Analysis and Machine Intelligence*, vol. 12, no. 5, pp. 447–458, 1990.

[18] A. P. Dempster, *A Generalization of Bayesian Inference*. Berlin, Heidelberg: Springer Berlin Heidelberg, 2008, pp. 73–104.

[19] M. Schreiber, V. Belagiannis, C. Gläser, and K. Dietmayer, "Dynamic occupancy grid mapping with recurrent neural networks," in *2021 IEEE International Conference on Robotics and Automation (ICRA)*, 2021, pp. 6717–6724.

[20] ——, "A multi-task recurrent neural network for end-to-end dynamic occupancy grid mapping," in *2022 IEEE Intelligent Vehicles Symposium (IV)*, 2022, pp. 315–322.

[21] K. Werber, M. Rapp, J. Klappstein, M. Hahn, J. Dickmann, K. Dietmayer, and C. Waldschmidt, "Automotive radar gridmap representations," in *2015 IEEE MTT-S International Conference on Microwaves for Intelligent Mobility (ICMIM)*, 2015, pp. 1–4.

[22] R. Prophet, H. Stark, M. Hoffmann, C. Sturm, and M. Vossiek, "Adaptions for automotive radar based occupancy gridmaps," in *2018 IEEE MTT-S International Conference on Microwaves for Intelligent Mobility (ICMIM)*, 2018, pp. 1–4.

[23] M. Li, Z. Feng, M. Stolz, M. Kunert, R. Henze, and F. Küçükay, "High resolution radar-based occupancy grid mapping and free space detection," in *Proceedings of the 4th International Conference on Vehicle Technology and Intelligent Transport Systems, VEHITS 2018, Funchal, Madeira, Portugal, March 16-18, 2018*. SciTePress, 2018, pp. 70–81.

[24] L. Sless, B. Shlomo, G. Cohen, and S. Oron, "Road scene understanding by occupancy grid learning from sparse radar clusters using semantic segmentation," 10 2019, pp. 867–875.

[25] R. Weston, S. Cen, P. Newman, and I. Posner, "Probably unknown: Deep inverse sensor modelling radar," in *2019 International Conference on Robotics and Automation (ICRA)*, 2019, pp. 5446–5452.

[26] J. E. Bresenham, "Algorithm for computer control of a digital plotter," *IBM Systems Journal*, vol. 4, no. 1, pp. 25–30, 1965.

[27] S. Thrun, W. Burgard, and D. Fox, *Probabilistic Robotics (Intelligent Robotics and Autonomous Agents)*. The MIT Press, 2005.

[28] L. Schmid, O. Andersson, A. Sulser, P. Pfreundschuh, and R. Siegwart, "Dynablox: Real-time detection of diverse dynamic objects in complex environments," vol. 8, no. 10, pp. 6259 – 6266, 2023.

[29] S. J. Steyer, "Grid-based object tracking," Ph.D. dissertation, Technische Universität München, 2021.

[30] L. McInnes, J. Healy, and S. Astels, "hdbscan: Hierarchical density based clustering," *The Journal of Open Source Software*, vol. 2, no. 11, p. 205, 2017.

[31] N. Scheiner, F. Kraus, N. Appenrodt, J. Dickmann, and B. Sick, "Object detection for automotive radar point clouds – a comparison," vol. 3, no. 6, 2021.

[32] F. Fent, P. Bauerschmidt, and M. Lienkamp, "Radargnn: Transformation invariant graph neural network for radar-based perception," in *Proceedings of the IEEE/CVF Conference on Computer Vision and Pattern Recognition (CVPR) Workshops*, June 2023, pp. 182–191.

[33] H. Caesar, V. Bankiti, A. H. Lang, S. Vora, V. E. Liong, Q. Xu, A. Krishnan, Y. Pan, G. Baldan, and O. Beijbom, "nuscenes: A multimodal dataset for autonomous driving," *arXiv preprint arXiv:1903.11027*, 2019.

[34] Z. Liu, H. Tang, A. Amini, X. Yang, H. Mao, D. Rus, and S. Han, "Bevfusion: Multi-task multi-sensor fusion with unified bird's-eye view representation," in *IEEE International Conference on Robotics and Automation (ICRA)*, 2023.